\documentclass{article}
\pdfoutput=1

\usepackage{spconf}
\usepackage{amsmath}
\usepackage{amsfonts}
\usepackage{graphicx}
\usepackage[numbers]{natbib}
\usepackage{hyperref}
\usepackage{algorithm}
\usepackage{algpseudocode}
\usepackage{multirow}
\usepackage{booktabs}
\usepackage{capt-of}

\title{GaLe: memory-efficient Global Approximate and Local Exact features}

\name{Alberto Ancilotto \qquad Elisabetta Farella \thanks{This work was supported by Ministero delle Imprese e del Made in Italy (IPCEI Cloud DM 27 giugno 2022 -- IPCEI-CL-0000007) and European Union (Next Generation EU).}}
\address{Fondazione Bruno Kessler (FBK), Trento, Italy}

\begin{document}
\maketitle

\begin{abstract}
Embedded devices typically lack the resources of GPU-equipped machines, and existing inference methods suffer from either high computational overhead (patch-based) or accuracy loss (approximation-based). We propose GaLe, a memory-efficient technique that enables the deployment of pretrained networks on constrained devices without retraining. GaLe partitions feature maps into two components: a local exact ($L_E$) representation that preserves fine details and a global approximate ($G_A$) representation that retains long-range dependencies. Unlike standard tiling, GaLe supports global operations and attention mechanisms found in hybrid CNN-transformer models. Validated on ImageNet, our method matches exact-inference performance while achieving up to $65\%$ speedup and $90\%$ RAM reduction on a \texttt{Cortex-M33} compared to patch-based inference. We further demonstrate GaLe's versatility across classification, detection, and generation tasks, highlighting its potential as a foundation for resource-efficient architecture design.
\end{abstract}


\section{Introduction}
Deep neural networks have grown increasingly resource-demanding, with modern attention-based architectures pushing the limits of available hardware. This is particularly true in embedded and TinyML systems, where the benefits of edge processing (in terms of privacy, latency, and energy efficiency) clash with strict hardware limitations. Typical MCU constraints include static memory (FLASH) restricting parameters to under 2MB, compute capacities orders of magnitude lower than GPUs, and most critically, dynamic memory (RAM) often below 1MB. RAM capacity is the primary bottleneck for modern building blocks like inverted residuals~\citep{howard2017mobilenets, sandler2018mobilenetv2, howard2019searching} and attention mechanisms~\citep{efficientvitMIT, dosovitskiy2020vits}. To address this challenge, we propose a novel memory-efficient inference strategy enabling the deployment of pre-trained networks on resource-constrained devices without retraining, requiring only a small calibration set. Our method relies on \textit{feature map partitioning}, where layer outputs are decomposed into two complementary representations: (i) a \emph{local exact ($L_E$)} component for fine-grained details and (ii) a \emph{global approximate ($G_A$)} component for long-range dependencies.
Unlike previous reduction methods~\citep{Lin2021MCUNetV2MP, patil2022poet}, this approach supports global receptive fields and attention blocks, minimizing accuracy loss while potentially guiding memory-aware architecture search. It also remains compatible with standard deployment runtimes such as ONNX, TFLite, and STM32Cube.Ai.
Our contributions are two-fold. First, we introduce a hardware-aware partial-patch-based inference method that optimizes memory usage via contiguous RAM slices. We reduce computational overhead by estimating the minimal patch overlap required to achieve target precision and by leveraging $G_A$ features to compensate for missing data. Second, we extend this approach to hybrid CNN-transformer architectures. We derive an attention formulation based on $G_A$ and $L_E$ features that enables sequential block processing with lower overhead than standard memory-efficient attention implementations~\citep{rabe2021memeffattn, dao2022flashattention}, and can operate alongside or replace them in memory-constrained settings.

\begin{figure}[tbp]
  \centering
   \includegraphics[width=.9\linewidth]{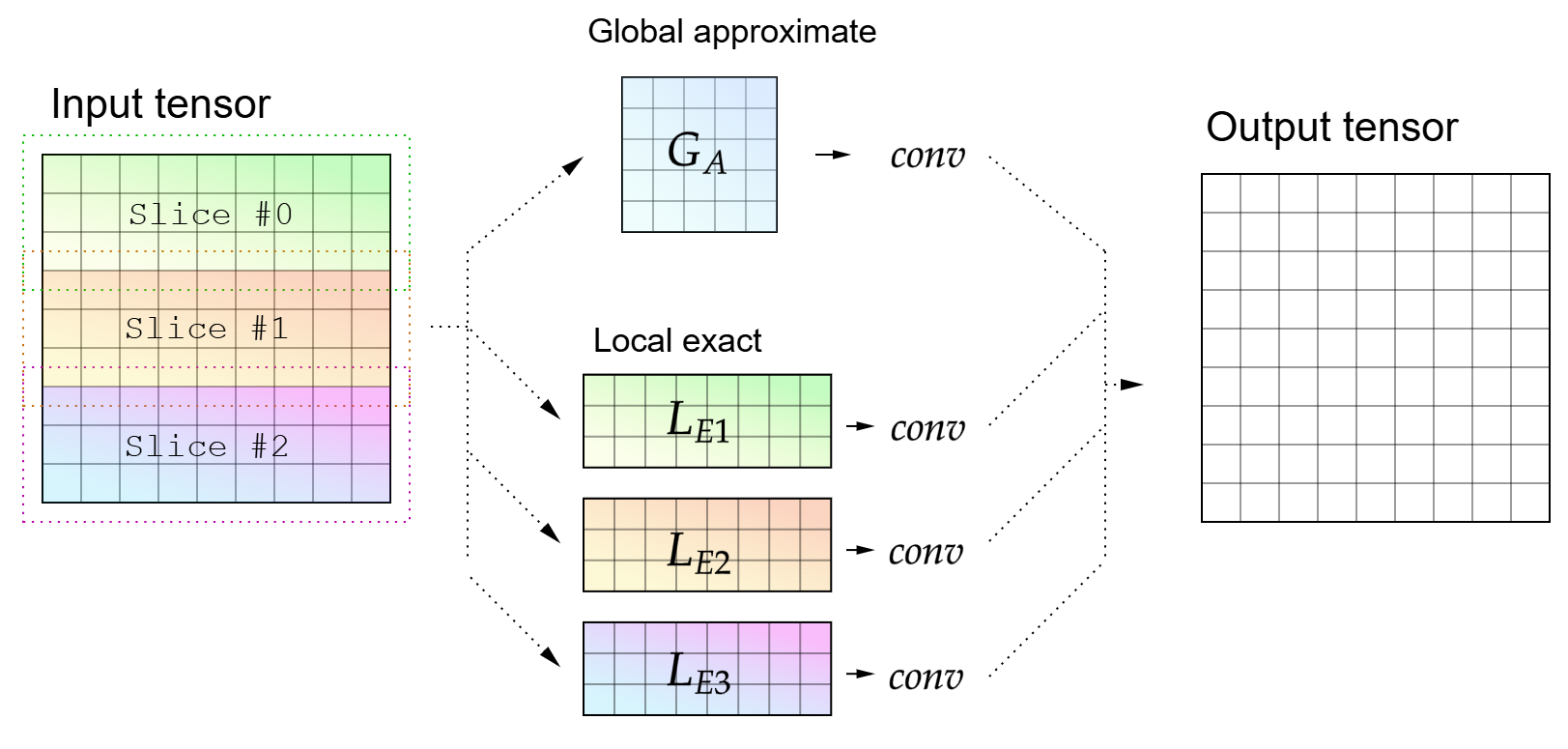}

   \caption{Graphical representation of \emph{local exact} and \emph{global approximate} decomposition for a convolutional layer}
   \label{fig:ga_le_approx}
\end{figure}

\section{Related works}
\label{sec:related}

\textbf{Training-based} approaches optimize network structure to reduce computational complexity. Foundational efficient architectures, such as the MobileNet family~\citep{ sandler2018mobilenetv2, howard2019searching, Qin2024MobileNetV4U} and Xception~\citep{chollet2017xception}, leverage depth-wise separable convolutions and inverted residual blocks. These concepts evolved into hardware-aware Neural Architecture Search (NAS) frameworks targeting MCUs, exemplified by MCUNet~\citep{lin2020mcunet, Lin2021MCUNetV2MP, lin2022mcunet3}, and scaling strategies like EfficientNet~\citep{tan2019efficientnet}, PhiNets~\citep{phinets}, and XiNets~\citep{Ancilotto_2023_ICCV}, which jointly optimize memory, parameters, and latency. Recent efforts extend these principles to efficient transformer designs~\citep{liu2021swin, Mehta2022SeparableSF} and fine-tuning based pruning strategies~\citep{wei2023joint, kim2022learned, wang2024multi}.
\textbf{Training-free} approaches reduce the memory usage of pre-trained models at runtime. The simplest method, reducing input resolution~\citep{tan2019efficientnet}, lowers RAM usage quadratically but degrades fine-grained details. Full patch-based inference~\citep{akyon2022slicing} preserves resolution but sacrifices global context. Partial patching, as adopted in TinyEngine~\citep{Lin2021MCUNetV2MP} and other embedded engines~\citep{GAPflow}, achieves mathematically exact results by tiling only the early layers; however, it is incompatible with global receptive field operators such as Squeeze-and-Excitation~\citep{hu2018squeeze} or attention. For transformers, runtime memory reduction often relies on token merging~\citep{bolya2022token, kim2024token} or exact algorithms like FlashAttention~\citep{dao2022flashattention, rabe2021memeffattn}, although the latter can incur significant computational overhead when aggressive memory reduction is required, limiting their applicability to embedded platforms.

\section{GaLe for Convolutional networks}
\label{sec:gale_conv_intro}
Classical memory-reduction approaches like partial patch-based inference (PPBI) suffer from high computational overhead~\citep{Lin2021MCUNetV2MP} and incompatibility with global operations like Squeeze-and-Excite (SE) modules~\citep{hu2018squeeze} or attention. To address these limitations, we propose a compute-efficient feature map slicing approach that approximates outputs using two complementary representations (Figure~\ref{fig:ga_le_approx}): a \textbf{Local Exact ($L_E$)} component, preserving full-resolution details for a spatial sub-region, and a \textbf{Global Approximate ($G_A$)} component, capturing low-resolution context over the entire map.
Unlike traditional patching, our method supports hybrid vision transformers and global receptive field operators while drastically reducing computational overhead (from over $100\%$ to less than $20\%$) through a hardware-aware tensor layout that optimizes cache locality.

\begin{figure}[t]
\centering
    \includegraphics[width=\linewidth]{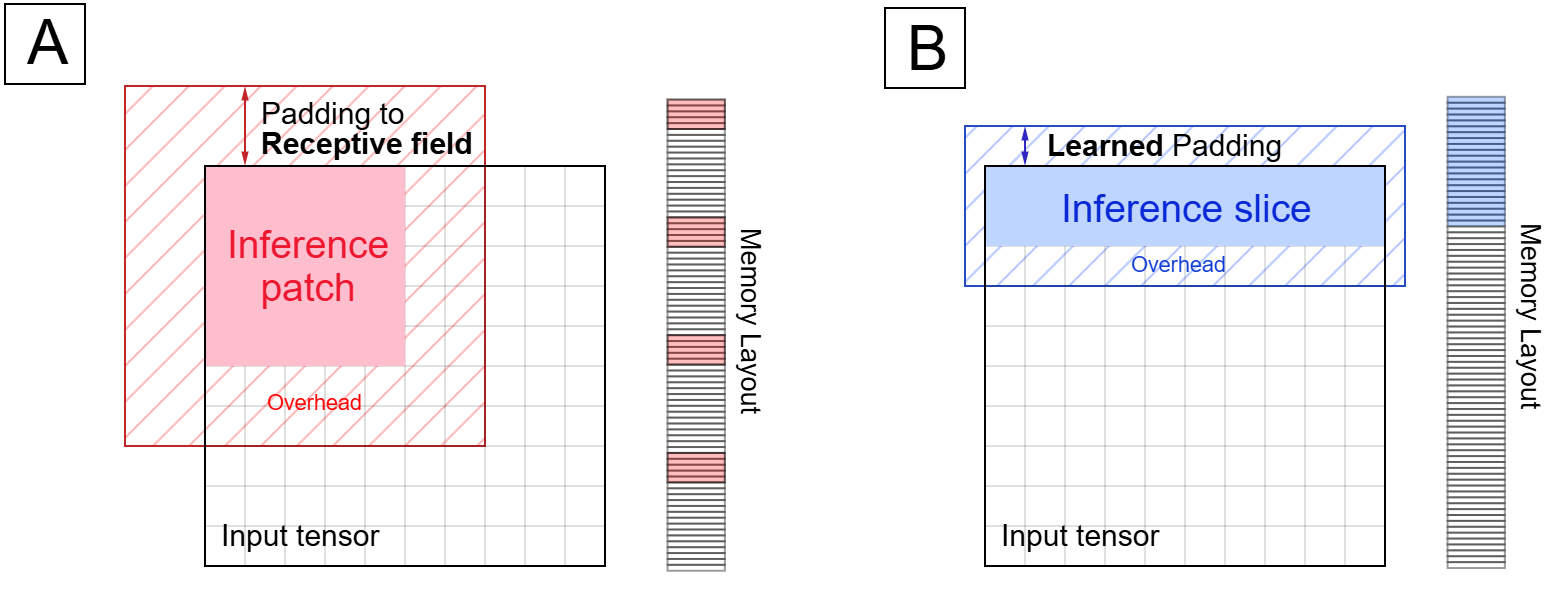}
    \caption{Proposed slicing vs classical PPBI. Our learned padding results in significantly lower overhead than padding to the full receptive field. Moreover, for \textit{NHWC} tensor inference, the proposed method relies on data stored in a single continuous memory area, thereby optimizing cache usage.}
    \label{fig:patch_vs_strip}
\end{figure}

\subsection{Computing Local Exact feature maps}
\label{sec:gale_conv_le}
In partial patch-based inference, the input is divided into $N$ smaller, partially overlapping areas, with spatial dimensions $P_H \times P_W$ that are processed independently - reducing the size of intermediate activation tensors. To ensure accurate outputs at patch boundaries, patches must overlap sufficiently so that the \emph{receptive field} of the layer at each output position is fully covered by valid input data. The required overlap $\mathcal{O}$ between patches is then equal to the receptive field of the deepest layer considered for patch-based execution: $\mathcal{O}_{ppbi} = \mathcal{R}_0(L)$, where $L$ is the layer index and $\mathcal{R}_0$ maps each layer to its receptive field on the input (layer $0$). For example, considering a CNN composed only of convolutional layers with kernel size $k_i$, stride $s_i$ for layer $i$, receptive field size $\mathcal{R}_0(L)$  on the input layer \citep{araujo2019computing} can be computed as:

\begin{equation}
\label{eq:rec_field}
    \mathcal{R}_0(L) = \sum_{l=1}^L \bigg((k_l-1) \prod_{i=1}^{l - 1} s_i \bigg) +1
\end{equation}

However, the receptive field size from Equation \ref{eq:rec_field} is only smaller than the input size for networks composed exclusively of local operations such as convolutional blocks and max-pooling layers. As such, PPBI can only reduce RAM requirements of a limited number of existing networks. In fact, this approach is not directly applicable to architectures that aggregate information across the entire input or operations with very large receptive fields, such as spatial pyramid pooling. Our approach generalizes partial patch-based inference to all network architectures by reducing the required overlap so that $\mathcal{O}_{GaLe}\leq \mathcal{O}_{ppbi}$, allowing for a user-controllable approximation error $\epsilon$ between the original and reconstructed feature maps. We rely on a \emph{calibration pass} to determine the appropriate overlap $\mathcal{O}_i$ for each network layer patch. During calibration, for each layer $i$, we iteratively increase the overlap $\mathcal{O}_i$ by $\mathcal{R}_i(i+1)$ if layer $i+1$ is a convolutional layer, or $s_{i+1}$ otherwise. This process continues until either the MSE between the reconstructed output tensor and the original output tensor falls below $\epsilon$ or the overlap reaches the maximum allowed based on memory constraints. In this case, we increase the number $N$ of subdivisions for the current layer.  Thanks to this iterative approximation strategy, the proposed method can approximate arbitrary network operators that may not be compatible with PPBI. When applied instead to architectures that can already be patched using traditional methods, GaLe can achieve significantly lower computational overhead, even for very small values of $\epsilon$. The proposed method is based on dividing the input tensor into horizontal \emph{slices} composed of multiple full rows ($P_W = W$) instead of square patches, as shown in Figure \ref{fig:patch_vs_strip}.  When working with tensors in NHWC format (the default for the majority of embedded runtimes \citep{tflite, cubeai}), slices align naturally with memory layout as shown in Figure \ref{fig:patch_vs_strip}. Each slice consists of values stored contiguously in memory, reducing cache misses and improving cache efficiency, resulting in lower latency, as shown in Section \ref{sec:results}.

\begin{figure}[tbp]
  \centering
  \includegraphics[width=\linewidth]{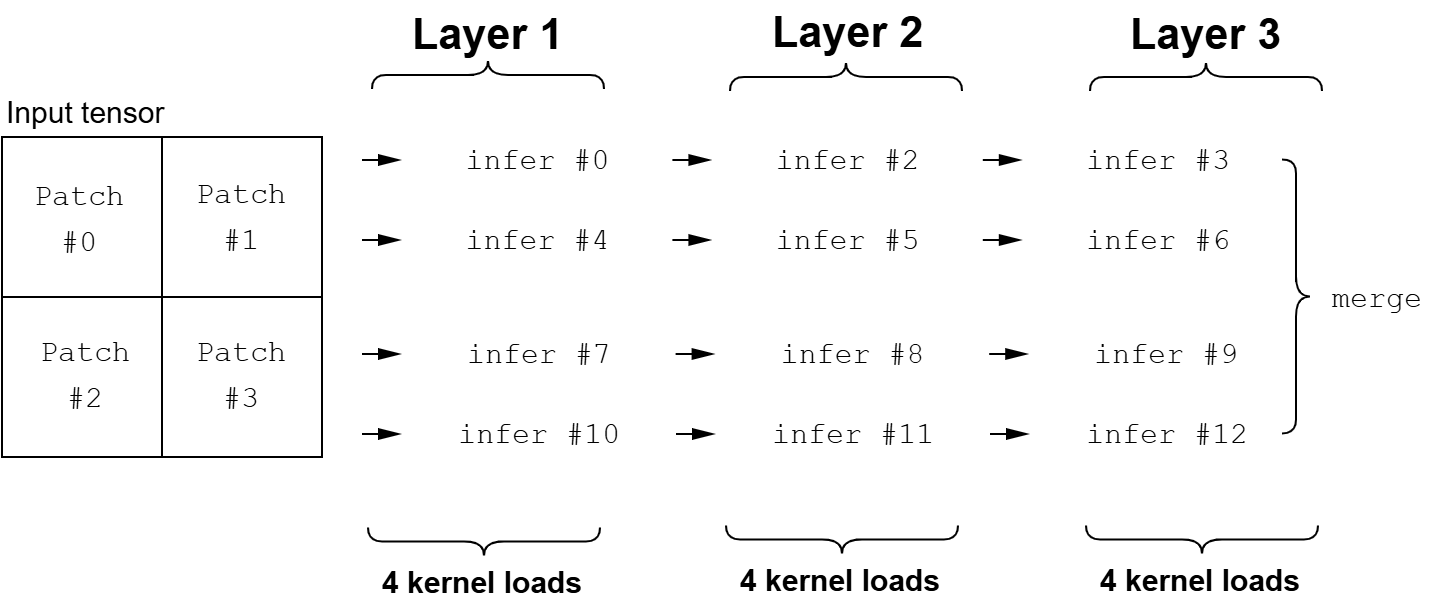}
   \includegraphics[width=\linewidth]{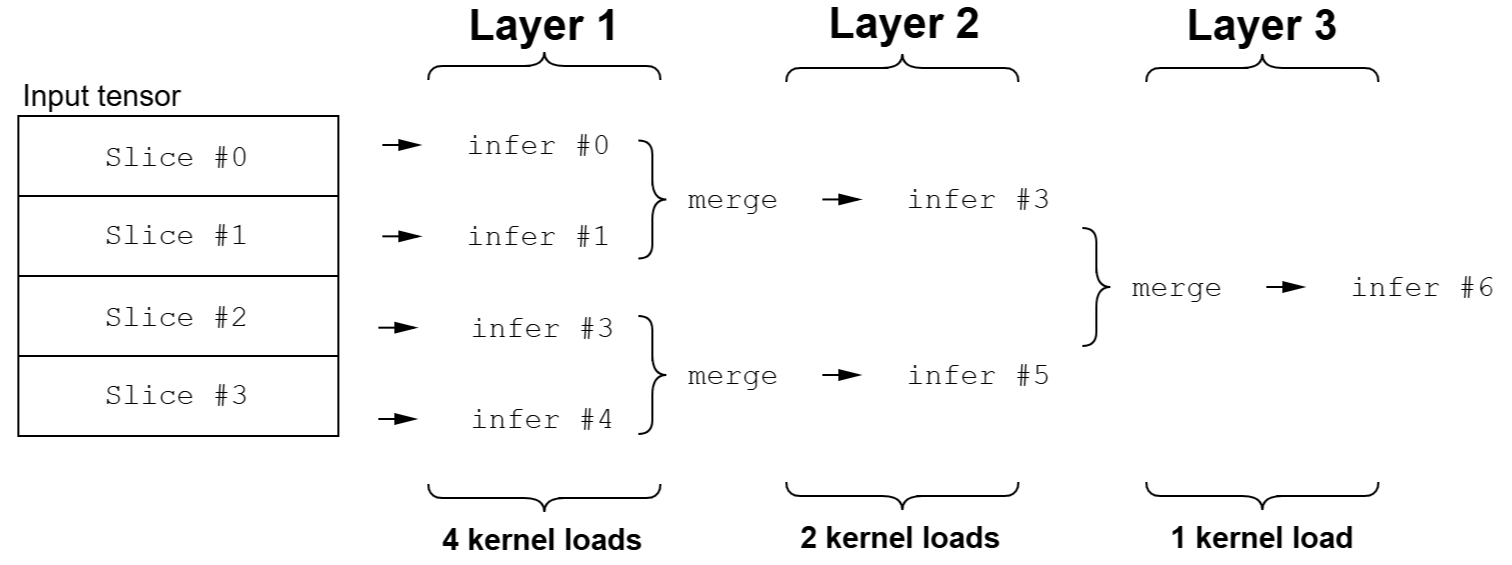}
   \caption{(Top) Example of partial patch-based inference on 3 layers with 4 patches resulting in 12 kernel loading operations. (Bottom) Slice-based inference on 3 layers with 4 patches, resulting in 6 kernel loading operations}
   \label{fig:kernel_load_slicing}
\end{figure}

\subsection{Optimizing memory transfers via Adaptive Slicing}

Standard PPBI enforces a fixed patch count across all layers, ignoring the natural reduction in spatial dimensions typical of convolutional backbones. This creates unnecessary fragmentation in deeper layers, leading to redundant computation due to patch overlaps and excessive memory traffic. To address this, GaLe employs \emph{adaptive slicing}. During calibration, we compute the minimal number of patches required for each block to strictly satisfy the peak memory constraint. As feature map resolution decreases, GaLe progressively reduces the patch count, merging slices whenever possible. This dynamic strategy minimizes the total overlap area, directly reducing redundant operations. Furthermore, by processing larger contiguous feature map sections, we drastically reduce the frequency of weight reloading from Flash to RAM (Figure \ref{fig:kernel_load_slicing}). Consequently, GaLe achieves significantly lower latency than fixed-pattern inference (Figure \ref{fig:latency_vs_ppbi}), particularly on MCUs where Flash bandwidth is a critical bottleneck.

\subsection{$G_A$ feature maps maintain global information}
\label{sec:gale_conv_ga}

Unlike standard PPBI, which relies on a large patch overlap to maintain mathematical equivalence, our $L_E$ patches are spatially isolated approximations that inherently lack global context. To address this, we introduce a \emph{Global Approximate ($G_A$)} branch. The $G_A$ input is generated by downsampling the original feature map by a factor $N$, which is determined during calibration to ensuring the entire reduced tensor fits within the target memory budget.
While various compression functions are possible, we find that simple bilinear downsampling offers the best trade-off between efficiency and representational quality. To construct the final output, we upsample the processed $G_A$ result and merge it with the fine-grained $L_E$ output using linear interpolation. Although computing the $G_A$ branch introduces a fixed overhead, it remains significantly faster than the overlapping required by PPBI (Figure \ref{fig:latency_vs_ppbi}), particularly as the reduction factor increases. Table \ref{tab:ablation} details the impact of combining $L_E$ and $G_A$ components on MobileNetV4 performance.

\begin{table}[t]
\centering
\begin{minipage}{0.3\linewidth}
\footnotesize
\begin{tabular}{lc}
\toprule
Feature Maps & Accuracy (\%) \\
\midrule
Ga + Le & 80.66 \\
Le only & 80.34 \\
Ga only & 79.16 \\
\bottomrule
\end{tabular}
\end{minipage}%
\hfill
\begin{minipage}{0.45\linewidth}
\captionof{table}{Comparison of feature map configurations on accuracy.}
\label{tab:ablation}
\end{minipage}
\end{table}

\section{GaLe for Hybrid Networks}
\label{sec:gale_attn_intro}
We extend the approach to modern attention-based and hybrid architectures, such as EfficientViT \citep{efficientvitMIT}, FastViT \citep{vasu2023fastvit} and MobileNetV4 \citep{Qin2024MobileNetV4U}.
As in the convolutional case, our method can offer superior performance than approximation-based techniques, with a fraction of the overhead required by math exact approaches based on memory-efficient attention (e.g. FlashAttention \cite{dao2023flashattention}), which recompute parts of the attention map multiple times during the forward pass.  Our technique proves particularly effective for hybrid architectures. In fact, unlike existing approaches, GaLe can be used to reduce the memory of both the convolutional and attention-based parts in a unified framework. By analyzing RAM usage patterns in these models, we tailor GaLe to enable deployment on ultra-low-resource devices such as MCUs, typically unsupported by existing methods.

\subsection{Attention via $L_E$ and $G_A$ Approximation}
\label{sec:gale_attn}
Standard scaled dot-product attention computes an output $\mathbf{O}$ from query, key, and value matrices $Q, K, V \in \mathbb{R}^{N \times d}$ as:
\begin{equation}
\label{eq:attn}
    \mathbf{O} = \text{Softmax}\left( \frac{QK^T}{\sqrt{d}} \right) V.
\end{equation}
This operation requires $\mathcal{O}(N^2)$ memory to store the attention map $\mathbf{A} = QK^T$. To mitigate this bottleneck without the overhead of recomputation-based methods~\cite{rabe2021memeffattn}, we propose an approximation scheme based on structured sparsity and low-rank components. We model the attention map $\mathbf{A}$ as a tiling of $b \times b$ blocks $\mathbf{A}_{ij}$, where each block is approximated as a weighted combination of a local sparse component and a global low-rank component:
\begin{equation}
    \mathbf{A}_{ij} \approx \alpha \mathbf{L}_{ij} + (1-\alpha) \mathbf{G}_{ij}.
\end{equation}
Here, $\mathbf{L}_{ij} = v_{ij} \mathbf{I}_b$ represents the fine-grained local correlation (where $\mathbf{I}_b$ is the identity matrix, preserving only diagonal dependencies within the block), and $\mathbf{G}_{ij} = c_{ij} \mathbf{J}_b$ represents the coarse global context (where $\mathbf{J}_b$ is the all-ones matrix of size $b \times b$). Transforming this structural approximation into a computable operation, we decouple the attention mechanism into $b$ independent ``strided" heads (Local Exact) and one ``downsampled" head (Global Approximate). This allows us to approximate the full operation on $N$ tokens as a weighted sum of operations on smaller subsets of size $N/b$.

\subsubsection{Algorithm Formulation}
Let the input sequences be partitioned into $b$ interleaved subsets (strides). We define the Local Exact attention term, which captures dependencies between tokens at the same phase $k \in \{1, \dots, b\}$, and the Global Approximate term, which captures dependencies on downsampled features:

\begin{equation}
\label{eq:attn_approx}
\begin{split}
    \widetilde{\text{Attn}}(Q,K,V) &= \alpha \sum_{k=1}^{b} \mathcal{F}\left( Q^{(k)}, K^{(k)}, V^{(k)} \right) \\ & + (1-\alpha) \mathcal{F}\left( Q_{ds}, K_{ds}, V_{ds} \right),
\end{split}
\end{equation}
where:
\begin{itemize}
    \item $\mathcal{F}(\cdot)$ denotes the standard Attention operation (Eq.~\ref{eq:attn});
    \item $Q^{(k)}, K^{(k)}, V^{(k)} \in \mathbb{R}^{\frac{N}{b} \times d}$ are the $k$-th strided slices of the inputs (e.g., $Q^{(k)}$ contains rows $k, k+b, k+2b \dots$);
    \item $Q_{ds}, K_{ds}, V_{ds} \in \mathbb{R}^{\frac{N}{b} \times d}$ are the inputs downsampled by factor $b$  to capture global context.
\end{itemize}

\subsubsection{Complexity Analysis}
By decomposing the problem, we avoid materializing the full $N \times N$ attention matrix. Instead, we compute $b+1$ smaller attention maps, each of size $\frac{N}{b} \times \frac{N}{b}$. This reduces the peak dynamic memory requirement for the attention map from $\mathcal{O}(N^2)$ to $\mathcal{O}((N/b)^2)$. Furthermore, because the operations in Eq.~\ref{eq:attn_approx} are independent, they can be executed sequentially, ensuring that the peak RAM usage remains bounded by the requirements of a single subset.

\begin{figure}[tbp]
\centering
    \includegraphics[width=.9\linewidth]{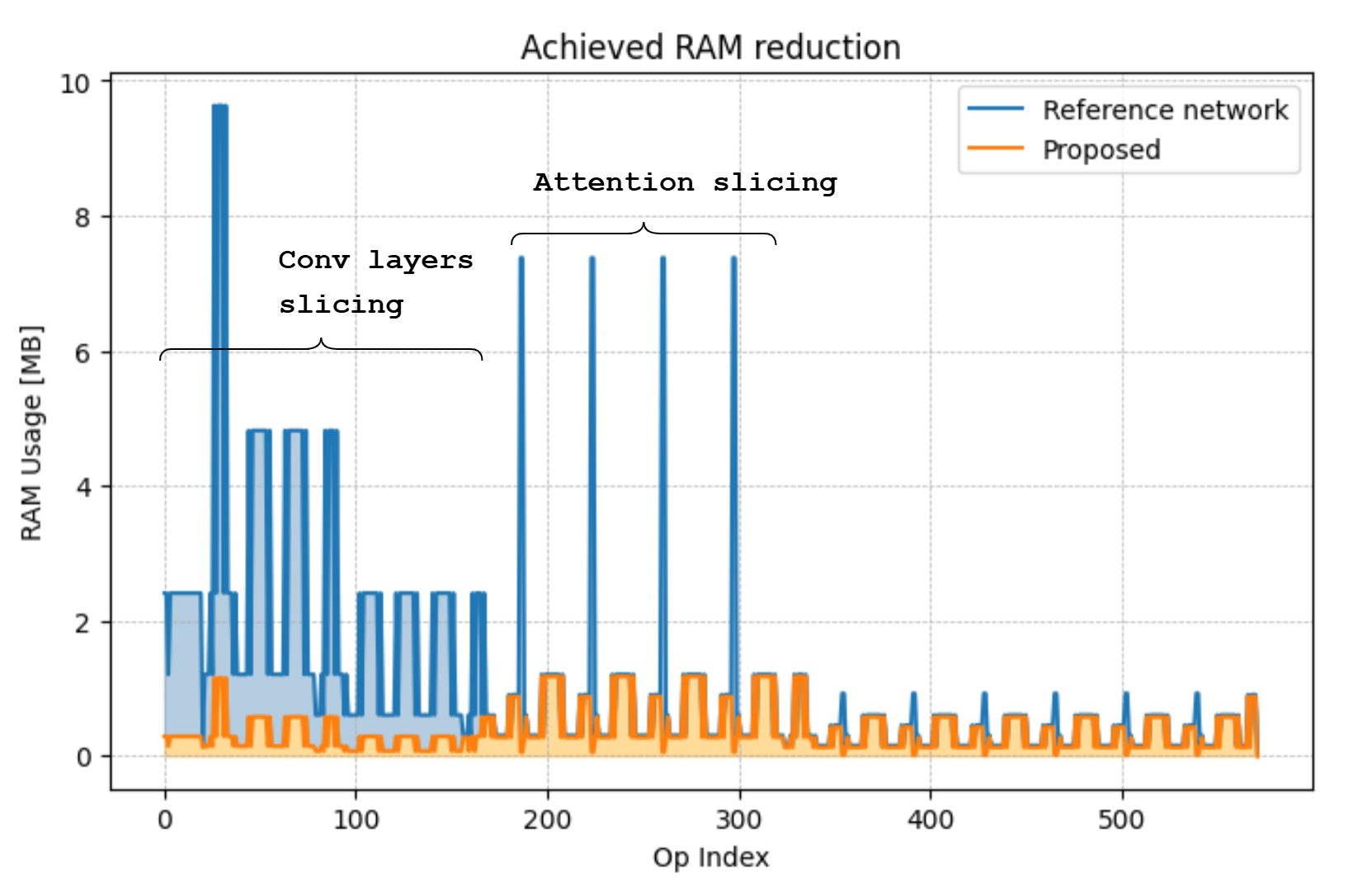}
    \caption{RAM requirements for EfficientVit-B2 \citep{efficientvitMIT} hybrid architecture, and savings achieved through the proposed approach. The contribution of conv blocks and attention to peak RAM usage are clearly seen.}
    \label{fig:ram_hybrid_archs}
\end{figure}

\subsubsection{Hybrid Architectures}
\label{sec:gale_hybrid_intro}

State-of-the-art efficient architectures often adopt hybrid designs, integrating convolutional layers for local feature extraction in early stages and attention mechanisms for global context in later stages~\citep{Qin2024MobileNetV4U, vasu2023fastvit, efficientvitMIT}.  As shown in Figure~\ref{fig:ram_hybrid_archs}, this structural heterogeneity creates a bi-modal memory profile, where peak RAM usage is governed by two distinct bottlenecks: Early Convolutional layers require memory to store the high-resolution activation maps, while attention operations generate peaks when the $\mathcal{O(HW^2)}$ attention map is computed.

\section{Results}
\label{sec:results}
\begin{figure}[h]
  \centering
   \includegraphics[width=.8\linewidth]{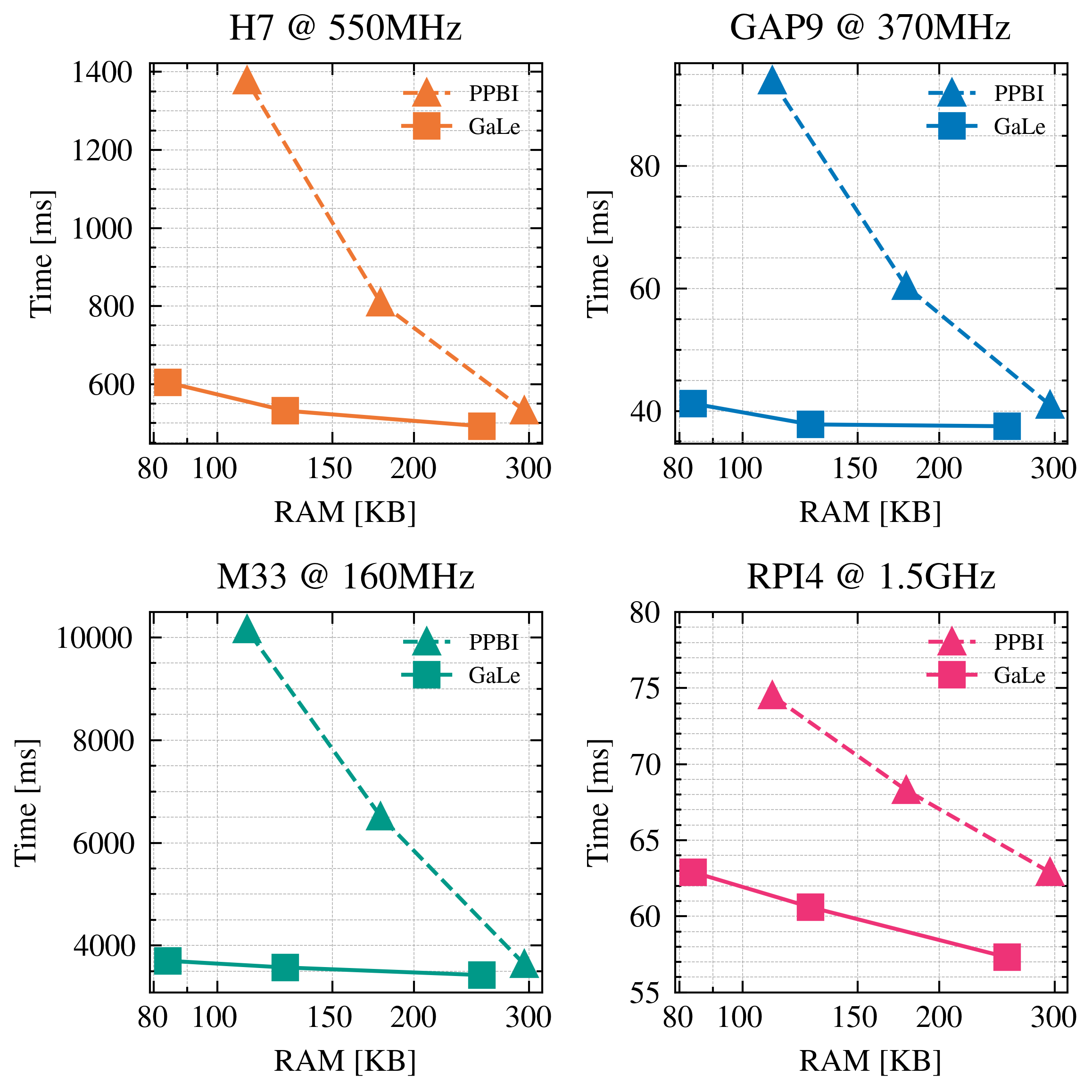}
   \caption{Latency vs RAM for \texttt{MobileNetV2} (256px, Cortex-M33). GaLe enables significant RAM savings at lower latencies than PPBI, achieving a $65\%$ speedup for a $90\%$ RAM reduction compared to tiling.}
   \label{fig:latency_vs_ppbi}
\end{figure}

We benchmarked GaLe on different hardware platforms, ranging from a \texttt{Raspberry Pi4} and \texttt{STM32U585} to the resource-constrained \texttt{STM32H743} and \texttt{GAP9}. Comparing our approach against resolution scaling and patch-based inference (FPBI/PPBI), we observe that GaLe consistently achieves superior memory reduction with minimal accuracy loss ($<1\%$) and significantly lower overhead (Figure \ref{fig:latency_vs_ppbi}, Table \ref{tab:results_imagenet}). Our results highlight specific architectural limitations of existing methods. While PPBI achieves an $88\%$ RAM reduction on MobileNetV2, it incurs an $80\%$ computational overhead; GaLe improves this reduction to $92\%$ with only $18\%$ overhead. Furthermore, GaLe succeeds where PPBI is structurally inapplicable: it handles the global dependencies of MobileNetV3's Squeeze-and-Excitation blocks via $G_A$ features, whereas standard tiling fails. Finally, for hybrid models like MobileNetV4 and FastViT, GaLe effectively compresses the entire model by addressing both convolutional and attention layers, where PPBI and Token Merging (ToMe) individually prove insufficient. Additional results for diffusion models are available in the appendix.

\begin{table*}[t]
\centering
\footnotesize
\begin{minipage}{0.55\linewidth}
\centering
\begin{tabular}{l l c c c}
\toprule
Network &  & RAM [KB] & Top1\% & Overhead\% \\
\midrule
\multirow{5}{*}{MobileNetV2 a110} & - & 6110 & 76.44 & - \\
& Res & 1527 & 64.16 & -74.7 \\
& PBI & 1588 & 66.72 & +2.1 \\
& PPBI & 763 & 76.44 & +78.3 \\
& \textbf{GaLe} & \textbf{476} & 76.42 & +18.7 \\
\midrule
\multirow{5}{*}{MobileNetV3 a100} & - & 4069 & 76.94 & - \\
& Res & 1017 & 62.50 & -73.9 \\
& FPBI & 1017 & 66.04 & +7.3 \\
& PPBI & / & / & / \\
& \textbf{GaLe} & \textbf{317} & 76.56 & +16.3 \\
\midrule
\multirow{5}{*}{ResNet50} & - & 9010 & 80.46 & - \\
& Res & 1216 & 66.94 & -85.8 \\
& FPBI & 1261 & 71.96 & +5.1 \\
& PPBI & 1171 & 80.46 & +167.9 \\
& \textbf{GaLe} & \textbf{847} & 79.52 & +40.9 \\
\bottomrule
\end{tabular}
\end{minipage}
\begin{minipage}{0.4\linewidth}
\centering
\begin{tabular}{l l c c c}
\toprule
Network &  & RAM [KB] & Top1\% & Overhead\% \\
\midrule
\multirow{6}{*}{MobileNetV4 HM} & - & 6314 & 81.80 & - \\
& Res & 1577 & 68.21 & -74.9 \\
& FPBI & 1623 & 69.08 & +1.6 \\
& ToMe & 6314 & 81.60 & -5.1 \\
& PPBI & 4922 & 81.81 & +26.4 \\
& \textbf{GaLe} & \textbf{884} & 80.66 & +14.6 \\
\midrule
\multirow{6}{*}{FastViT} & - & 8342 & 81.96 & - \\
& Res & 1042 & 65.70 & -79.7 \\
& FPBI & 2085 & 74.60 & +5.4 \\
& ToMe & 8342 & 81.80 & -9.2 \\
& PPBI & 1668 & 81.96 & +43.4 \\
& \textbf{GaLe} & \textbf{992} & 81.10 & +5.9 \\
\midrule
\multirow{6}{*}{EfficientViT} & - & 9732 & 82.46 & - \\
& Res & 1430 & 68.72 & -83.4 \\
& PPBI & 7687 & 82.46 & +31.4 \\
& ToMe & 9732 & 82.26 & -6.1 \\
& FPBI & 1625 & 72.21 & +4.1 \\
& \textbf{GaLe} & \textbf{1294} & 81.48 & +15.6 \\
\bottomrule
\end{tabular}
\end{minipage}
\caption{Comparison of GaLe against state-of-the-art methods on ImageNet. GaLe achieves the best trade-off between RAM reduction and computational overhead across all architectures.}
\label{tab:results_imagenet}
\end{table*}

\subsection{Case study: Object Detection}
\label{sec:case_study_obj_det}
\begin{figure}[h]
  \centering
   \includegraphics[width=.7\linewidth]{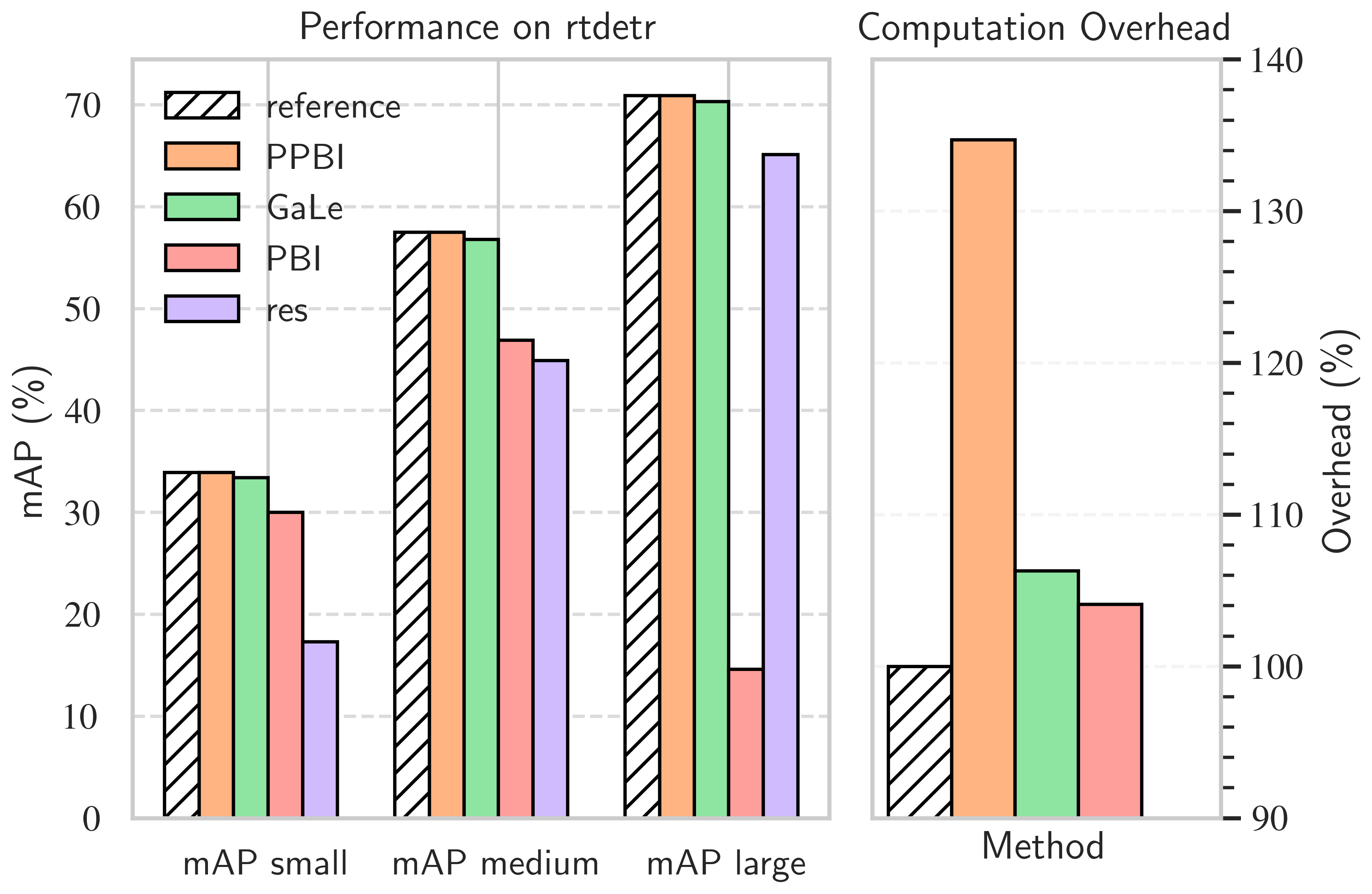}
  \hspace{.05\linewidth}
   \includegraphics[width=.7\linewidth]{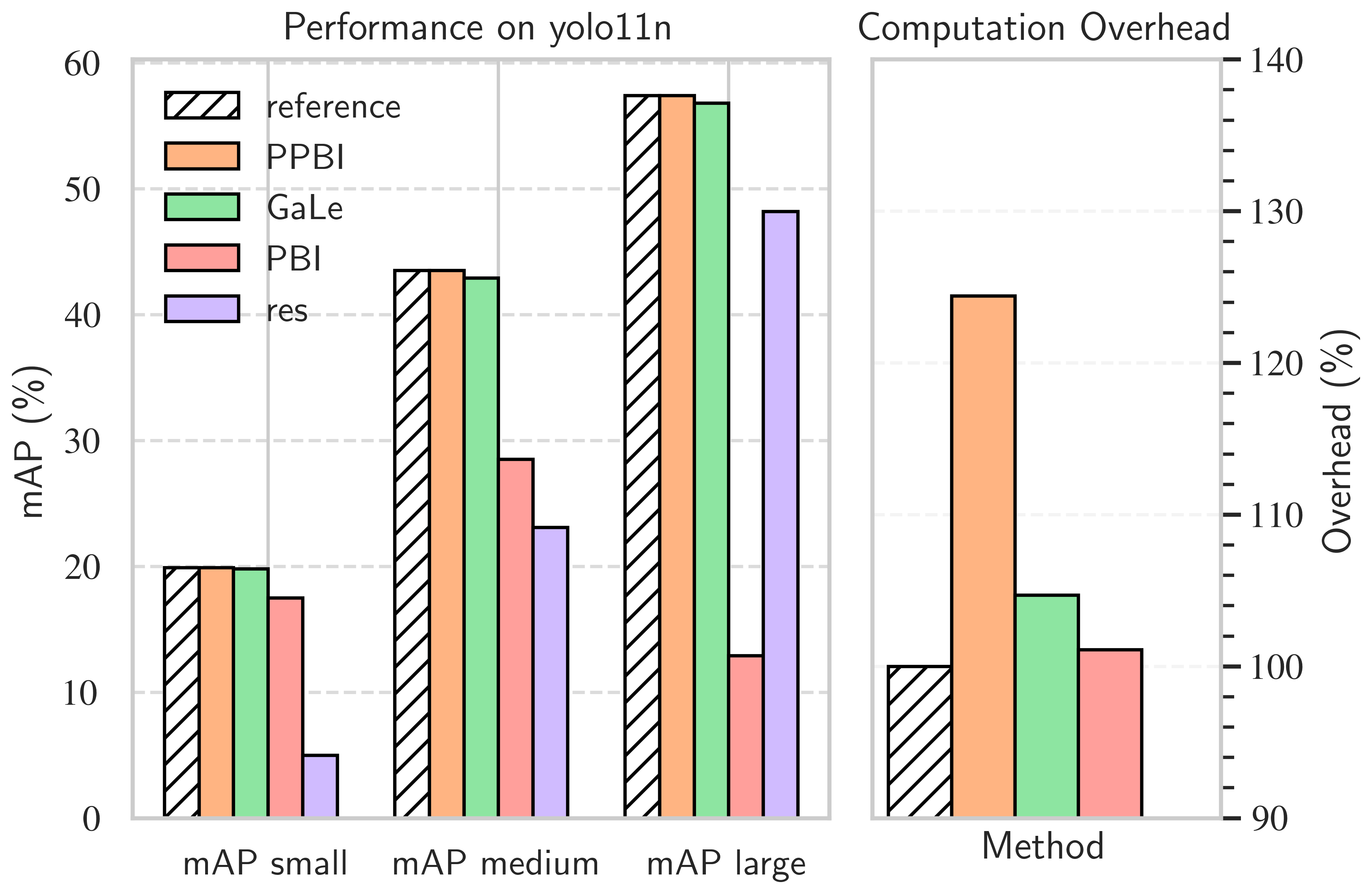}
   \caption{Performance comparison for RT-DETR-L (target $<2.5$MB) and YOLOv11n (target $<512$KB). GaLe maintains mAP comparable to exact methods with minimal overhead.}
   \label{fig:obj_det_yolo}
\end{figure}

We evaluated GaLe on object detection, deploying quantized RT-DETR-L (requiring 9.8MB RAM) and YOLOv11n (4.2MB RAM) onto constrained targets: a high-performance 2.5MB platform (e.g., GAP9, STM32H7) and a low-power 512KB platform (e.g., STM32L4). Figure \ref{fig:obj_det_yolo} illustrates the trade-offs. Resolution scaling degrades performance for small objects, while FPBI degrades performance for large objects due to spatial fragmentation. PPBI preserves accuracy but incurs in huge computational overhead. In contrast, GaLe effectively balances efficiency and accuracy, achieving negligible mAP reduction by maintaining both local detail and global context via the $L_E/G_A$ decomposition.

\section{Conclusion}
This work introduced GaLe, a feature map partitioning strategy enabling the deployment of pretrained deep learning models on memory-constrained devices without retraining. By decomposing activations into local exact and global approximate components, GaLe resolves the limitations of existing tiling methods, ensuring compatibility with global operators and attention mechanisms. Validation on ImageNet classification demonstrates similar performance with exact inference methods while delivering up to $90\%$ memory reduction and substantial speedups over patch-based methods. Further experiments on object detection and diffusion highlight GaLe's versatility, positioning it as a foundational technique for efficient edge AI and a potential driver for future memory-aware neural architecture search.

{
    \small
    \bibliographystyle{IEEEbib}
    \bibliography{refs}

\begin{thebibliography}{10}

\bibitem{howard2017mobilenets}
{Howard, Andrew G et al},
\newblock ``Mobilenets: Efficient convolutional neural networks for mobile
  vision applications,''
\newblock {\em arXiv preprint arXiv:1704.04861}, 2017.

\bibitem{sandler2018mobilenetv2}
{Sandler, Mark, et al},
\newblock ``Mobilenetv2: Inverted residuals and linear bottlenecks,''
\newblock in {\em Proceedings of the IEEE conference on computer vision and
  pattern recognition}, 2018.

\bibitem{howard2019searching}
Andrew Howard et~al.,
\newblock ``Searching for mobilenetv3,''
\newblock in {\em Proceedings of the IEEE/CVF international conference on
  computer vision}, 2019, pp. 1314--1324.

\bibitem{efficientvitMIT}
Han Cai et~al.,
\newblock ``Efficientvit: Lightweight multi-scale attention for high-resolution
  dense prediction,''
\newblock in {\em Proceedings of the IEEE/CVF International Conference on
  Computer Vision}, 2023, pp. 17302--17313.

\bibitem{dosovitskiy2020vits}
Alexey Dosovitskiy et~al.,
\newblock ``An image is worth 16x16 words: Transformers for image recognition
  at scale,''
\newblock in {\em International Conference on Learning Representations}, 2021.

\bibitem{Lin2021MCUNetV2MP}
Ji~Lin, Wei-Ming Chen, Han Cai, Chuang Gan, and Song Han,
\newblock ``Mcunetv2: Memory-efficient patch-based inference for tiny deep
  learning,''
\newblock in {\em Neural Information Processing Systems}, 2021.

\bibitem{patil2022poet}
Shishir Patil et~al.,
\newblock ``Poet: Training neural networks on tiny devices with integrated
  rematerialization and paging,''
\newblock in {\em International Conference on Machine Learning}. PMLR, 2022.

\bibitem{rabe2021memeffattn}
Markus~N Rabe and Charles Staats,
\newblock ``Self-attention does not need $o(n^2)$ memory,''
\newblock {\em arXiv preprint arXiv:2112.05682}, 2021.

\bibitem{dao2022flashattention}
Tri Dao, Dan Fu, Stefano Ermon, Atri Rudra, and Christopher R{\'e},
\newblock ``Flashattention: Fast and memory-efficient exact attention with
  io-awareness,''
\newblock {\em Advances in neural information processing systems}, vol. 35, pp.
  16344--16359, 2022.

\bibitem{Qin2024MobileNetV4U}
Danfeng Qin et~al.,
\newblock ``Mobilenetv4 - universal models for the mobile ecosystem,''
\newblock {\em European Conference on Computer Vision (ECCV)}, vol.
  abs/2404.10518, 2024.

\bibitem{chollet2017xception}
Fran{\c{c}}ois Chollet,
\newblock ``Xception: Deep learning with depthwise separable convolutions,''
\newblock in {\em Proceedings of the IEEE conference on computer vision and
  pattern recognition}, 2017, pp. 1251--1258.

\bibitem{lin2020mcunet}
Ji~Lin, Wei-Ming Chen, Yujun Lin, John Cohn, Chuang Gan, and Song Han,
\newblock ``Mcunet: Tiny deep learning on iot devices,''
\newblock {\em arXiv preprint arXiv:2007.10319}, 2020.

\bibitem{lin2022mcunet3}
Ji~Lin et~al.,
\newblock ``On-device training under 256kb memory,''
\newblock {\em Advances in Neural Information Processing Systems}, vol. 35, pp.
  22941--22954, 2022.

\bibitem{tan2019efficientnet}
Mingxing Tan and Quoc Le,
\newblock ``Efficientnet: Rethinking model scaling for convolutional neural
  networks,''
\newblock in {\em International Conference on Machine Learning}. PMLR, 2019.

\bibitem{phinets}
Francesco Paissan, Alberto Ancilotto, and Elisabetta Farella,
\newblock ``Phinets: A scalable backbone for low-power ai at the edge,''
\newblock {\em ACM Trans. Embed. Comput. Syst.}, feb 2022.

\bibitem{Ancilotto_2023_ICCV}
Alberto Ancilotto, Francesco Paissan, and Elisabetta Farella,
\newblock ``Xinet: Efficient neural networks for tinyml,''
\newblock in {\em Proceedings of the IEEE/CVF International Conference on
  Computer Vision (ICCV)}, October 2023, pp. 16968--16977.

\bibitem{liu2021swin}
Ze~Liu, Yutong Lin, Yue Cao, Han Hu, Yixuan Wei, Zheng Zhang, Stephen Lin, and
  Baining Guo,
\newblock ``Swin transformer: Hierarchical vision transformer using shifted
  windows,''
\newblock in {\em Proceedings of the IEEE/CVF international conference on
  computer vision}, 2021, pp. 10012--10022.

\bibitem{Mehta2022SeparableSF}
Sachin Mehta and Mohammad Rastegari,
\newblock ``Separable self-attention for mobile vision transformers,''
\newblock {\em Transactions on Machine Learning Research}, 2023.

\bibitem{wei2023joint}
Siyuan Wei et~al.,
\newblock ``Joint token pruning and squeezing towards more aggressive
  compression of vision transformers,''
\newblock in {\em Proceedings of the IEEE/CVF conference on computer vision and
  pattern recognition}, 2023, pp. 2092--2101.

\bibitem{kim2022learned}
Sehoon Kim et~al.,
\newblock ``Learned token pruning for transformers,''
\newblock in {\em Proceedings of the 28th ACM SIGKDD Conference on Knowledge
  Discovery and Data Mining}, 2022, pp. 784--794.

\bibitem{wang2024multi}
Yunke Wang, Bo~Du, Wenyuan Wang, and Chang Xu,
\newblock ``Multi-tailed vision transformer for efficient inference,''
\newblock {\em Neural Networks}, vol. 174, pp. 106235, 2024.

\bibitem{akyon2022slicing}
Fatih Akyon et~al.,
\newblock ``Slicing aided hyper inference and fine-tuning for small object
  detection,''
\newblock in {\em 2022 IEEE international conference on image processing
  (ICIP)}. IEEE, 2022, pp. 966--970.

\bibitem{GAPflow}
GreenWaves Technologies,
\newblock ``Automated design intelligence with gapflow,''
  \url{https://greenwaves-technologies.com/}.

\bibitem{hu2018squeeze}
Jie Hu, Li~Shen, and Gang Sun,
\newblock ``Squeeze-and-excitation networks,''
\newblock in {\em Proceedings of the IEEE conference on computer vision and
  pattern recognition}, 2018, pp. 7132--7141.

\bibitem{bolya2022token}
Daniel Bolya, Cheng-Yang Fu, Xiaoliang Dai, Peizhao Zhang, Christoph
  Feichtenhofer, and Judy Hoffman,
\newblock ``Token merging: Your vit but faster,''
\newblock {\em arXiv preprint arXiv:2210.09461}, 2022.

\bibitem{kim2024token}
Minchul Kim et~al.,
\newblock ``Token fusion: Bridging the gap between token pruning and token
  merging,''
\newblock in {\em Proceedings of the IEEE/CVF Winter Conference on Applications
  of Computer Vision}, 2024, pp. 1383--1392.

\bibitem{araujo2019computing}
Andr{\'e} Araujo, Wade Norris, and Jack Sim,
\newblock ``Computing receptive fields of convolutional neural networks,''
\newblock {\em Distill}, vol. 4, no. 11, pp. e21, 2019.

\bibitem{tflite}
google,
\newblock ``tflite-micro,'' \url{https://github.com/tensorflow/tflite-micro}.

\bibitem{cubeai}
ST~Microelectronics,
\newblock ``X-cube-ai,''
  \url{https://www.st.com/en/embedded-software/x-cube-ai.html}.

\bibitem{vasu2023fastvit}
Pavan Kumar~Anasosalu Vasu et~al.,
\newblock ``Fastvit: A fast hybrid vision transformer using structural
  reparameterization,''
\newblock in {\em ICCV}, 2023.

\bibitem{dao2023flashattention}
Tri Dao,
\newblock ``Flashattention-2: Faster attention with better parallelism and work
  partitioning,''
\newblock {\em arXiv preprint arXiv:2307.08691}, 2023.

\bibitem{sauer2024sdturbo}
Axel Sauer, Dominik Lorenz, Andreas Blattmann, and Robin Rombach,
\newblock ``Adversarial diffusion distillation,''
\newblock in {\em European Conference on Computer Vision}. Springer, 2024, pp.
  87--103.

\bibitem{johnson2016perceptual}
Justin Johnson, Alexandre Alahi, and Li~Fei-Fei,
\newblock ``Perceptual losses for real-time style transfer and
  super-resolution,''
\newblock in {\em Computer Vision--ECCV 2016: 14th European Conference,
  Amsterdam, The Netherlands, October 11-14, 2016, Proceedings, Part II 14}.
  Springer, 2016, pp. 694--711.

\end{thebibliography}
}

\appendix
\section{Appendix}
\subsection{Case study: diffusion models}
\label{sec:case_study_sd_turbo}
We present an additional application of GaLe in the context of diffusion models as a replacement for self-attention layers to reduce memory consumption while avoiding the computational overhead associated with memory-efficient attention techniques \citep{rabe2021memeffattn}. To evaluate the effectiveness of our approach, we benchmark SD-Turbo \citep{sauer2024sdturbo} comparing GaLe against memory-efficient attention and token merging (ToMe) for attention memory reduction. Additionally, we introduce a hybrid variant in which token merging is applied together with GaLe, computing $G_A$ through ToMe, while the $L_E$ features remain unchanged.

\begin{figure}[h]
  \centering
   \includegraphics[width=\linewidth]{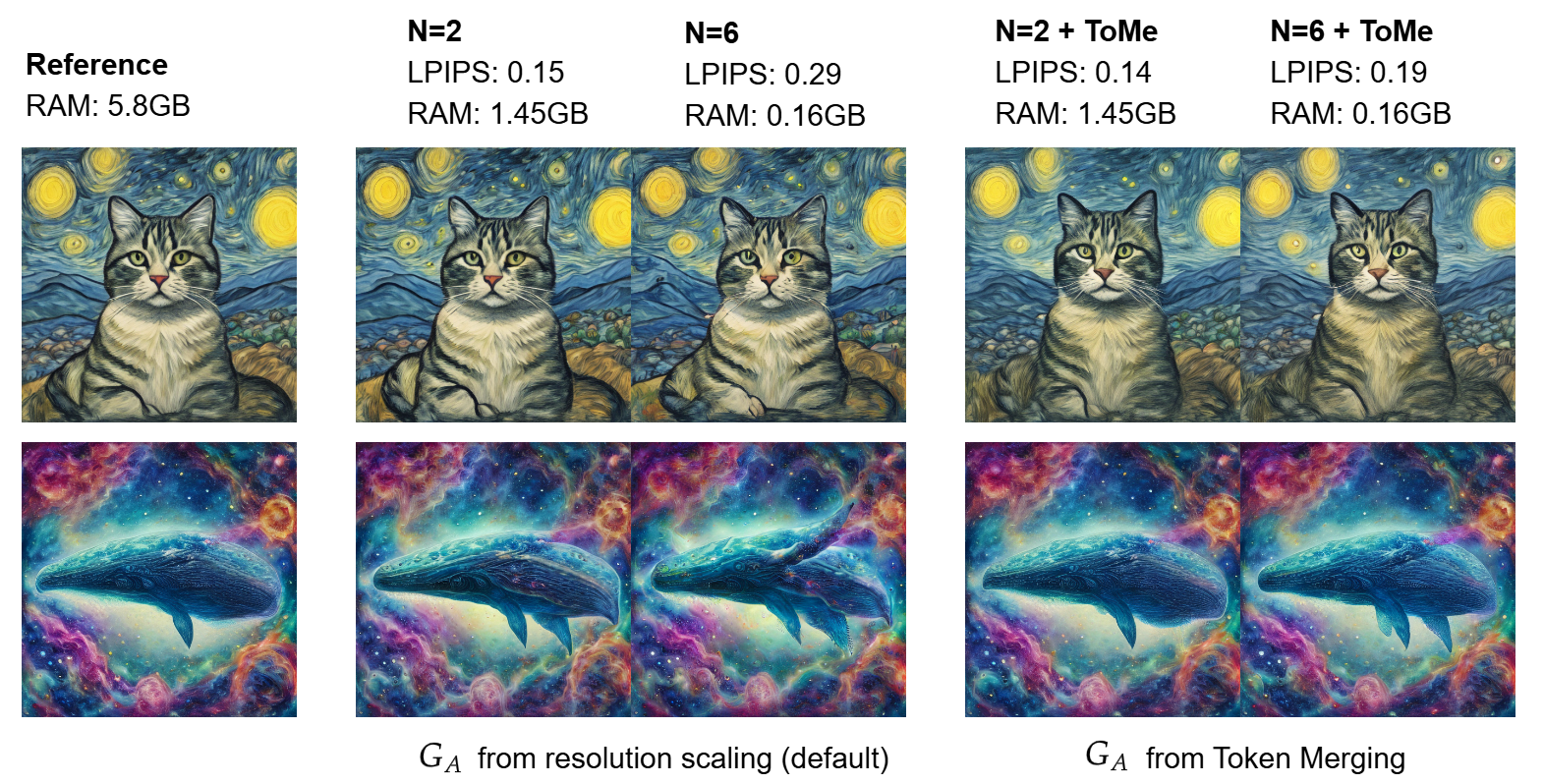}

   \caption{Our approach allows for significant RAM savings with minimal quality loss when applied to SD-turbo \citep{sauer2024sdturbo}. Token merging strategies can be used for computing the Ga feature maps, increasing performance.}
   \label{fig:sd_cats}
\end{figure}

\begin{figure}[tbp]
  \centering
   \includegraphics[width=.93\linewidth]{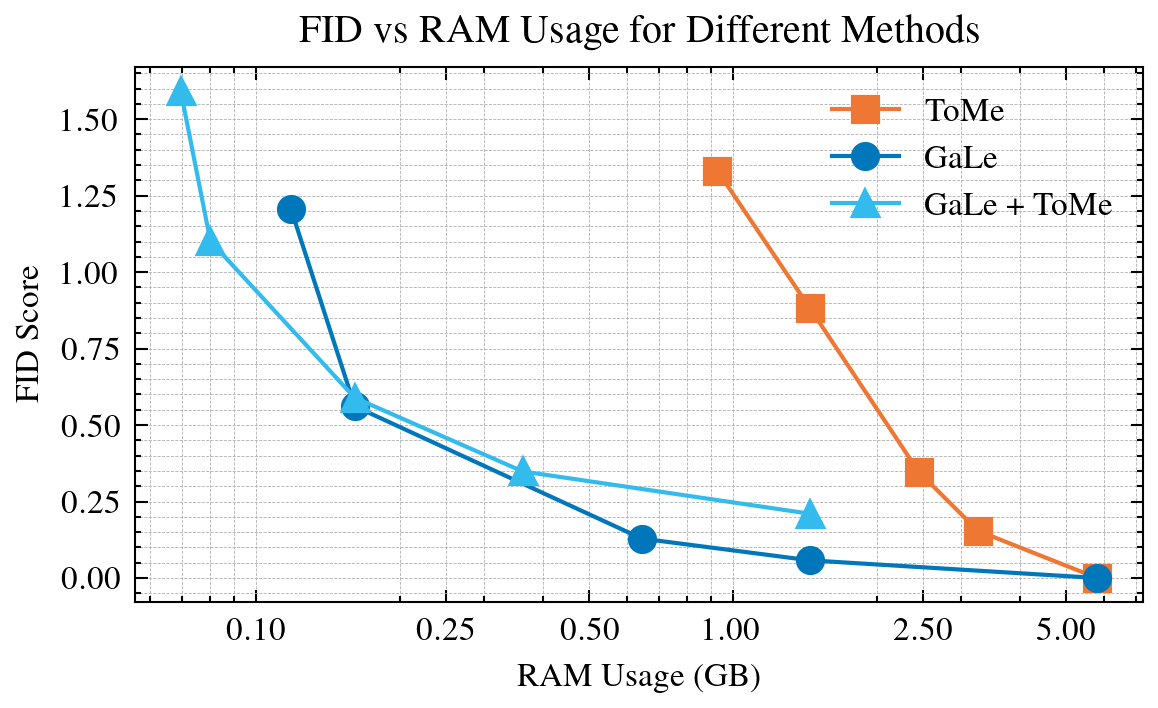}
   \includegraphics[width=.9\linewidth]{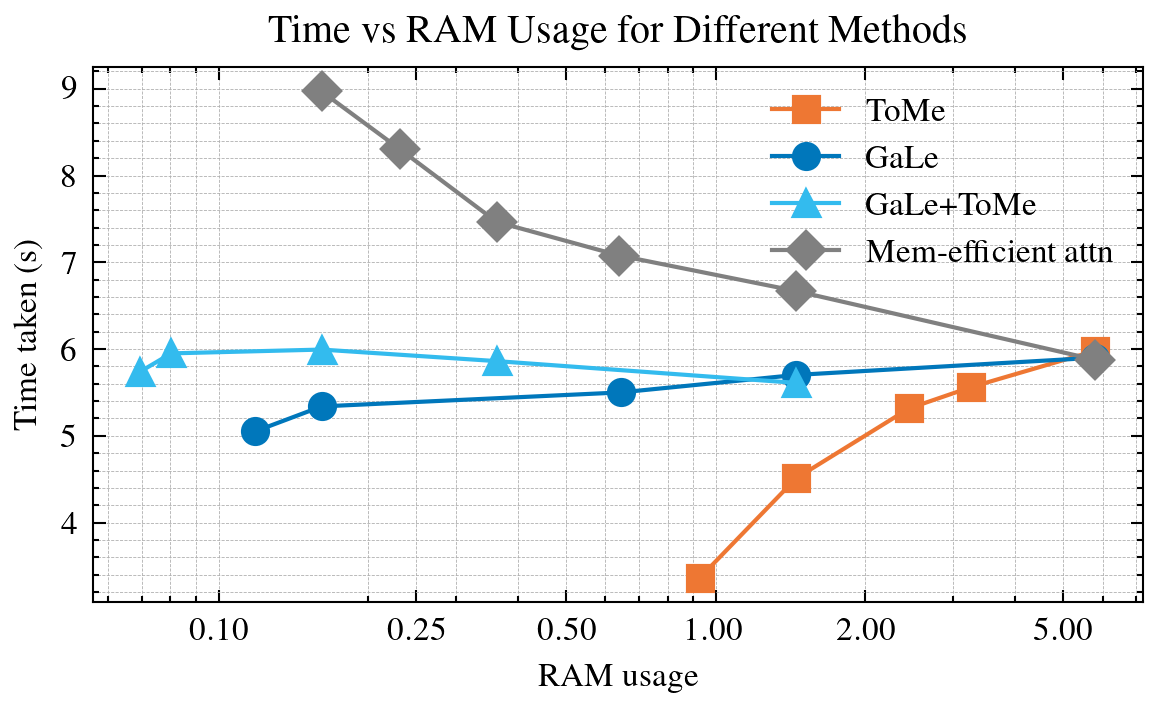}

   \caption{(Top) Time required for generating five $512\times512$ images on a mobile  RTX3000 Ada. GaLe allows for significantly reduced RAM usage without the computational overhead of using memory-efficient attention. (Bottom) Compared to token merging approaches, GaLe allows for significantly reduced RAM usage. Token merging can be used for computing the Ga feature map, allowing for lower FID.}
   \label{fig:fid_time_sd}
\end{figure}

\begin{figure}[tbp]
  \centering
   \includegraphics[width=.8\linewidth]{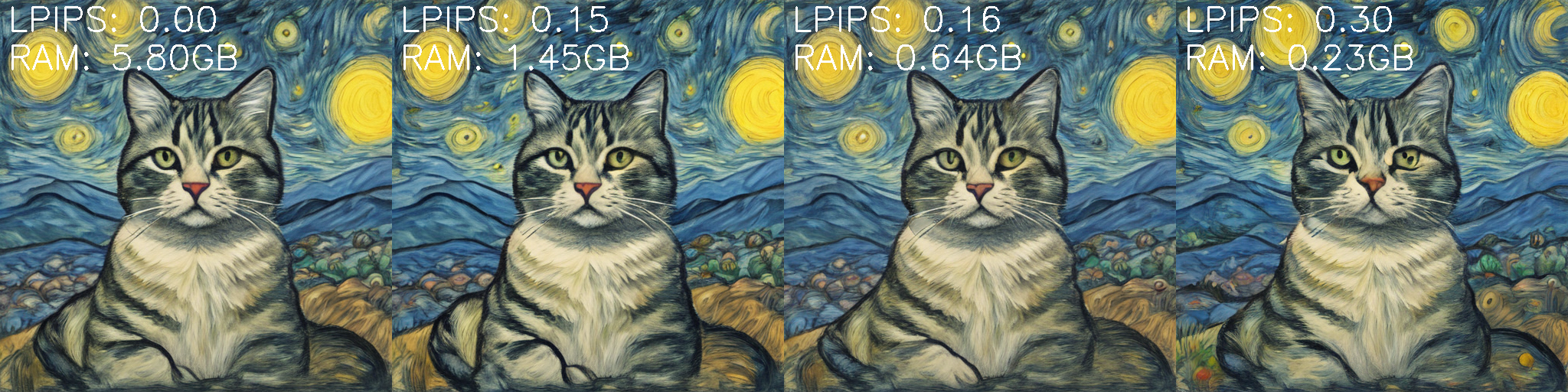}
   \includegraphics[width=.8\linewidth]{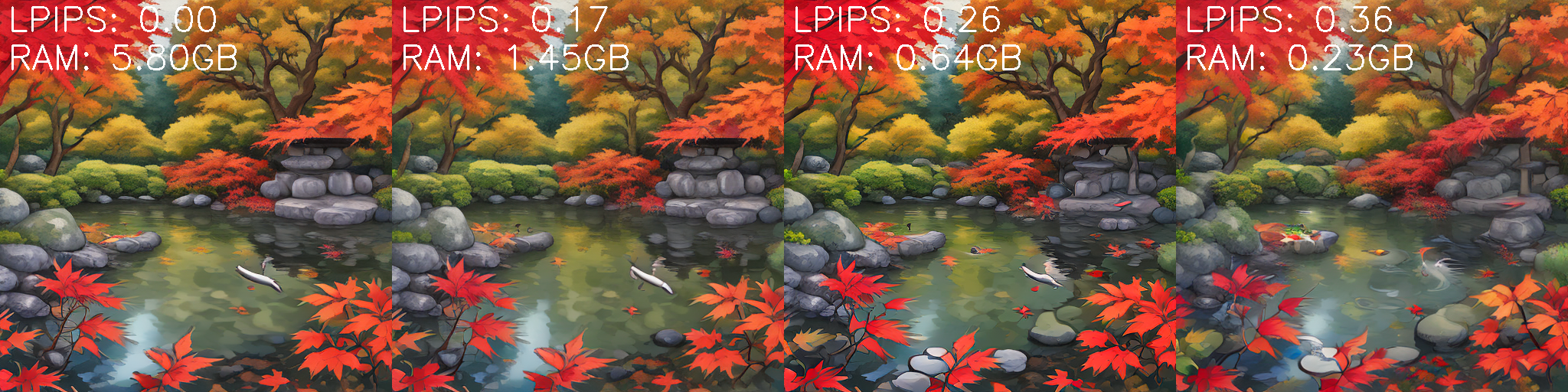}
   \includegraphics[width=.8\linewidth]{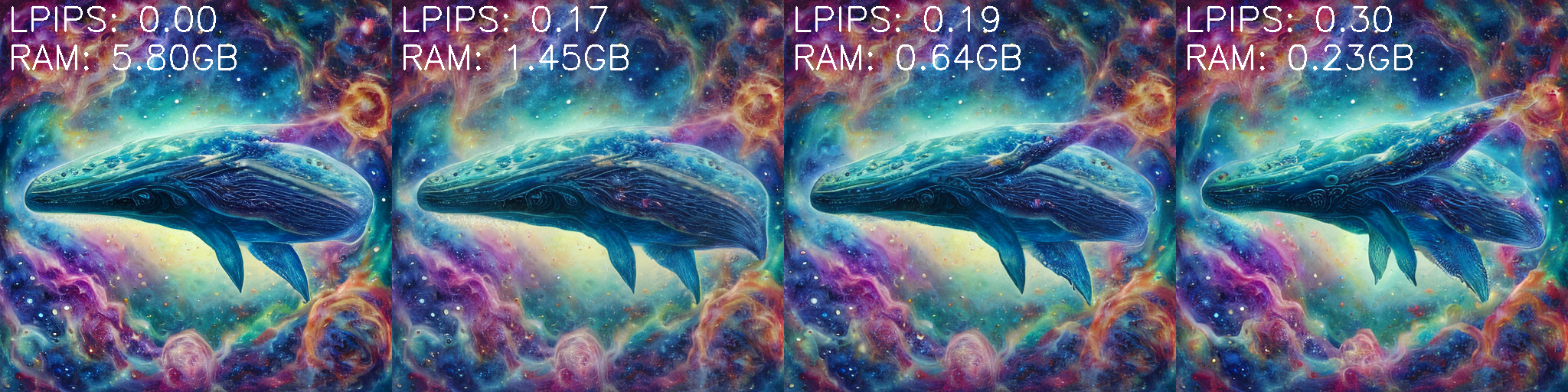}
   \includegraphics[width=.8\linewidth]{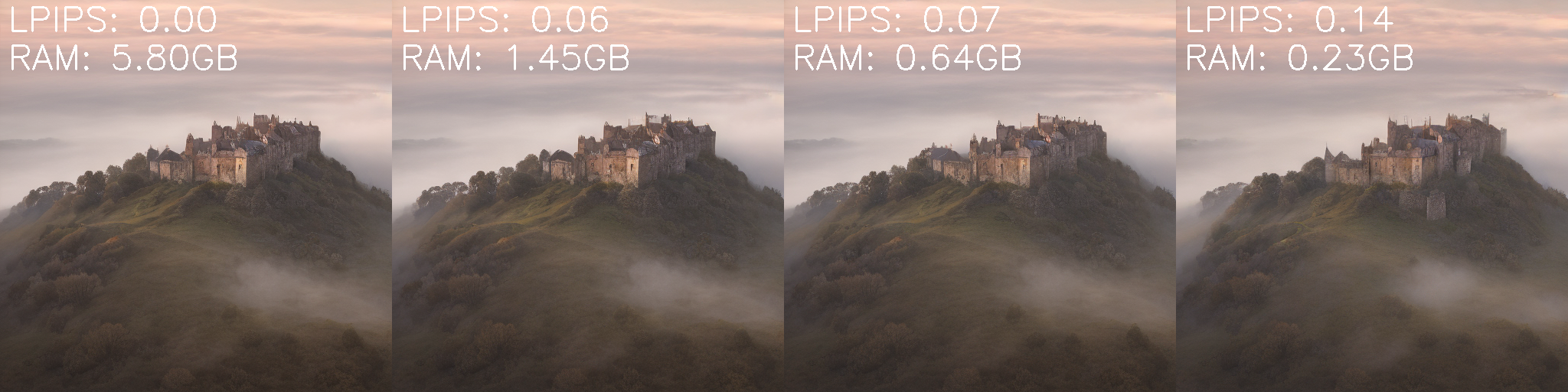}
   \includegraphics[width=.8\linewidth]{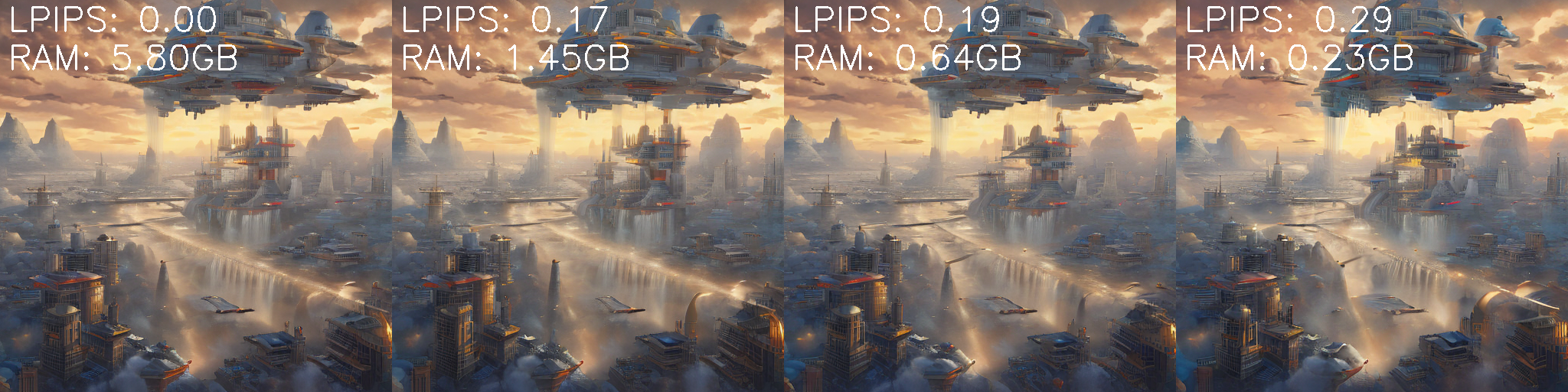}
   \includegraphics[width=.8\linewidth]{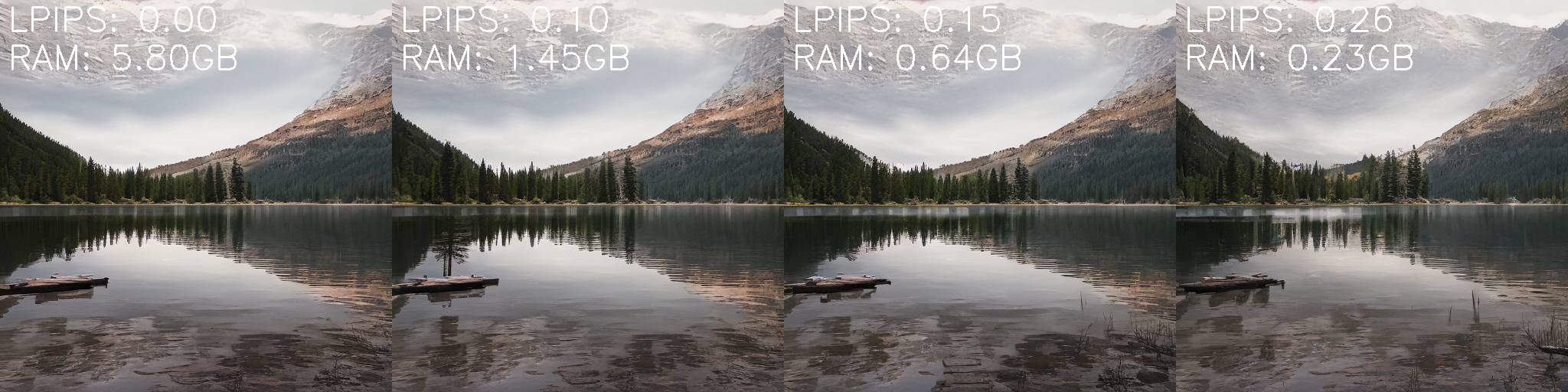}

   \caption{Example images using sd-turbo with GaLe.}
   \label{fig:more_sd_turbo}
\end{figure}

As shown in Figure \ref{fig:fid_time_sd}, GaLe achieves a substantially greater reduction in memory usage compared to ToMe alone, preserving the fidelity of generated images, as measured by Fr\'echet Inception Distance (FID) and Perceptual Similarity score (LPIPS \citep{johnson2016perceptual}) relative to the baseline network, while proving faster than standard memory-efficient attention. When GaLe is combined with ToMe, extreme memory compression factors can be achieved, which would otherwise be unattainable using ToMe alone, while maintaining a low FID score. Figure \ref{fig:more_sd_turbo} shows examples of images generated from SD-Turbo using GaLe, with the achieved RAM reductions and perceptual similarity scores. Table \ref{tab:high_res_generation} shows the performance obtained when generating images at different resolutions. We can see how the proposed approach offers high performance even at extreme (25$\times$) compression ratios both for lower (512$\times$512) and higher (1024$\times$1024) generation resolutions.

\subsection{Detailed platform performance}
Table \ref{tab:mbv2_a05} shows the real-world speedup achieved using GaLe compared to traditional partial-patch based inference. Higher speedups are obtained for simpler MCUs (eg \texttt{Cortex M33}), while more powerful devices with tiered caches (e.g. a \texttt{Raspberry Pi4}) do not benefit as much from the memory-aware slice layout.

\begin{table}[tbp]
    \centering
    \small
    \begin{tabular}{l c c c c c c}
        \toprule
         & N & RAM [KB] & H743 & GAP9 & M33 & RPi4 \\
        \midrule
        Base & 1 & 1016 & 432 & 37.08 & 3132 & -- \\
        \midrule
        PPBI & 4  & 295  & 532  & 40.8  & 3632  & 62.9 \\
        PPBI & 8  & 178  & 808  & 60.4  & 6512  & 68.3 \\
        PPBI & 16 & 111  & 1378 & 93.96 & 10159 & 74.5 \\
        \midrule
        GaLe & 4  & 254  & 492  & 37.48 & 3424  & 57.3 \\
        GaLe & 8  & 127  & 532  & 37.8  & 3572  & 60.6 \\
        GaLe & 16 & 84   & 604  & 41.2  & 3702  & 62.9 \\
        \bottomrule
    \end{tabular}

    \caption{Performance comparison of MBV2 A05 256x on different platforms. Times in ms}
    \label{tab:mbv2_a05}
\end{table}

\subsection{GaLe calibration pass}
The following pseudocode shows the calibration pass logic. Slice count is increased until the target performance are met. At each iteration, one forward pass is performed, the error is evaluated, and the overlap for the $i-th$ layer is increased.
\begin{algorithm}
\caption{Calibration of slice Overlap Parameters}
\label{alg:calibration}
\small
\begin{algorithmic}[1]
\Require Network, error tolerance $\epsilon$, total RAM
\Ensure Overlap values $\{O_i\}_{i=1}^L$

\For{$i = 1$ to $L$}
    \State $O_i \gets 0$
    \State $N \gets ceil(\frac{RAM_i}{total\  RAM})$
    \Repeat

    \Repeat
        \State Run N-Le-based inference, overlap $O_i$
        \State Compute $\mathrm{MSE}_i$ for output of layer $i$
        \If{Layer $i+1$ is a convolutional layer}
            \State $O_i \gets O_i + \mathcal{R}_i(i+1)$
        \Else
            \State $O_i \gets O_i + s_{i+1}$
        \EndIf
    \Until{$\mathrm{MSE}_i < \epsilon$ \textbf{or} memory limit exceeded}
    \State $N \gets N+1$
    \Until{\textbf{not} memory limit exceeded}

\EndFor

\end{algorithmic}
\end{algorithm}

\section{Generalization}
We report additional results in Table \ref{tab:inaturalist}, obtained by applying GaLe to a pretrained ViT on iNaturalist to evaluate fine-grained classification performance. We can see that the approach can maintain high performance even on this task, with significant RAM reductions and minimal performance loss, demonstrating generalization also when used with huge numbers of classes.

\begin{table}[tbp]
\centering
\begin{tabular}{ccc}
\toprule
Split & Accuracy & RAM Usage (M) \\
\midrule
1  & 91.98  & 6170.89 \\
2  & 91.797 & 3396.60 \\
4  & 90.732 & 1851.10 \\
16 & 89.859 & 519.27  \\
\bottomrule
\end{tabular}
\caption{Fine-grained Classification (iNaturalist) Results}
\label{tab:inaturalist}
\end{table}

\begin{table*}[tbp]
\centering
\begin{tabular}{ccccccc}
\toprule
Attn split & LPIPS (512x512) & RAM (GB) & LPIPS (768x768) & RAM (GB) & LPIPS (1024x1024) & RAM (GB) \\
\midrule
1  & 0.00 & 5.77 & 0.00 & 12.98 & 0.00 & 23.07 \\
4  & 0.15 & 1.44 & 0.14 & 3.24  & 0.13 & 5.77  \\
9  & 0.16 & 0.64 & 0.20 & 1.44  & 0.09 & 2.56  \\
25 & 0.30 & 0.23 & 0.21 & 0.52  & 0.15 & 0.92  \\
\bottomrule
\end{tabular}
\caption{Image Generation Results (High-resolution)}
\label{tab:high_res_generation}
\end{table*}

\section{Sensitivity Analysis of Weighting Factor $\alpha$}
\label{sec:alpha_sensitivity}

\begin{figure}[h]
    \centering
    \includegraphics[width=0.9\linewidth]{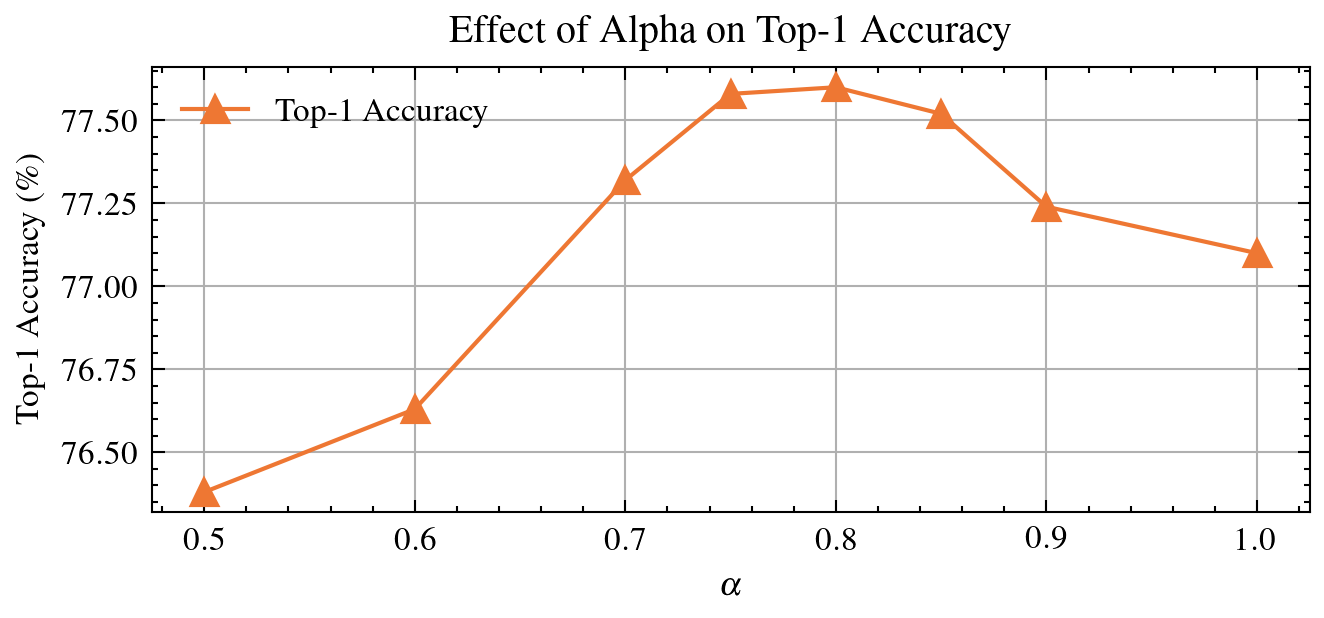}
    \caption{Effect of the weighting factor $\alpha$ on Top-1 Accuracy. The peak at $\alpha=0.8$ demonstrates that a hybrid representation outperforms both pure local slicing ($\alpha=1.0$) and aggressive global approximation ($\alpha=0.5$).}
    \label{fig:alpha_sensitivity}
\end{figure}

To validate the necessity of our hybrid decomposition strategy, we conducted a sensitivity analysis on the weighting factor $\alpha$, which controls the fusion between the Local Exact ($L_E$) and Global Approximate ($G_A$) feature maps. As defined in the method, the approximated output is given by a weighted combination where $\alpha$ governs the contribution of fine-grained local details versus global semantic context. Figure~\ref{fig:alpha_sensitivity} illustrates the impact of $\alpha$ on Top-1 Accuracy for ImageNet classification. Results indicate that the method is highly stable within the range $\alpha \in [0.7, 0.9]$, with peak performance observed at $\alpha \approx 0.8$. Notably, performance degrades at the extremes:
\begin{itemize}
    \item At lower values (e.g., $\alpha=0.5$), the model relies too heavily on the downsampled $G_A$ map, resulting in a loss of high-frequency details.
    \item At $\alpha=1.0$ (pure slicing), the accuracy drops ($\approx 77.1\%$) compared to the hybrid peak ($\approx 77.6\%$). This confirms that when memory constraints prevent full receptive field overlap, the $L_E$ component alone suffers from insufficient global context, which is effectively recovered by the $G_A$ component.
\end{itemize}

\section{Calibration Efficiency and Sample Size}
\label{sec:calibration_samples}

\begin{figure}[h]
    \centering
    \includegraphics[width=0.9\linewidth]{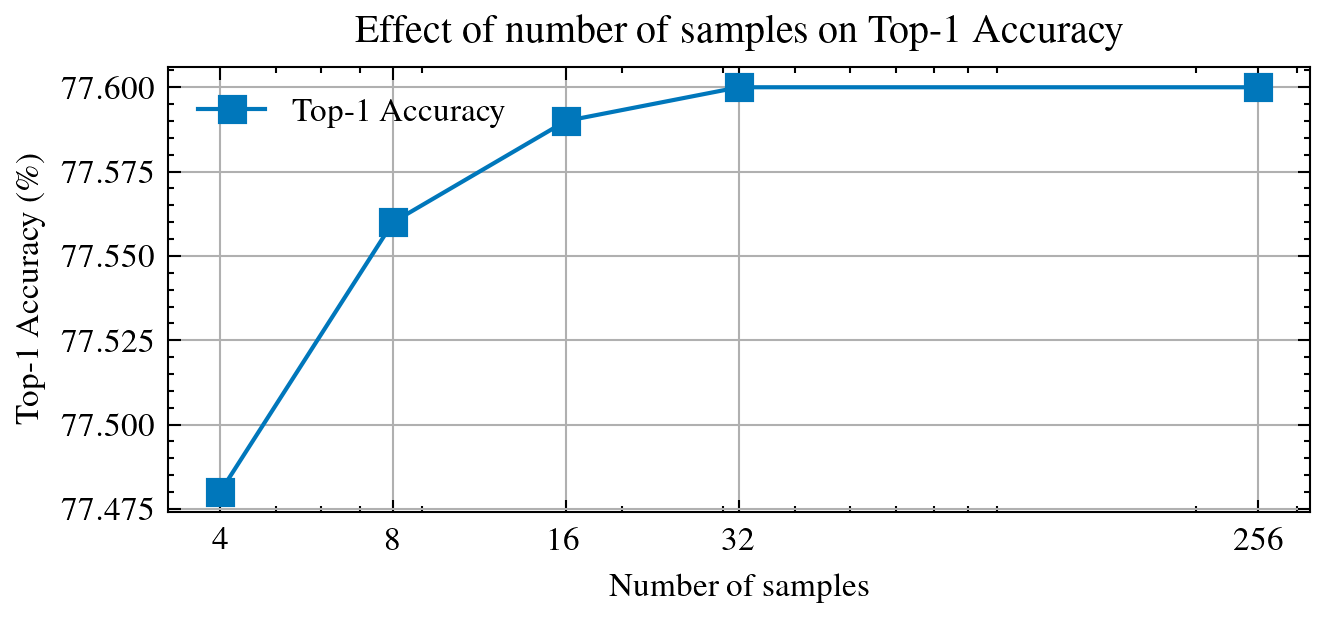}
    \caption{Effect of the number of calibration samples on Top-1 Accuracy. The method achieves optimal performance stability with as few as 32 samples, demonstrating high data efficiency and low setup overhead.}
    \label{fig:calibration_samples}
\end{figure}

We analyzed the relationship between the size of the calibration dataset and the resulting model accuracy to assess data efficiency.
As shown in Figure~\ref{fig:calibration_samples}, the calibration process exhibits rapid convergence. The Top-1 Accuracy stabilizes significantly with as few as 16 to 32 samples, with negligible performance gains observed when increasing the sample size to 256. This data efficiency implies that the calibration step incurs minimal computational overhead and can be performed rapidly offline before deployment, without requiring large-scale validation sets or extensive processing time.

\section{Additional results on Timm models}
\label{sec:appendix_timm}
Tables \ref{tab:additional_timm_1} and \ref{tab:additional_timm_2} show additional results achieved on common \textit{timm} models using GaLe. Configurations are identified with the starting slice number ($S$), the number of split blocks ($B$), the allowed MSE error for the calibration pass ($\epsilon$) and the feature maps are used (for small RAM reductions, $L_E$ features already provide the same performance as the full network, so no $G_A$ features are used to reduce overhead).

\begin{table*}[h]
\centering
\begin{tabular}{llccc}
\toprule
Network & Configuration & RAM Reduction \% & Top-1 \% & Overhead \% \\
\midrule
\multirow{5}{*}{MobilenetV2 a50}&
 - & - & 67.18 & - \\
  &
 2S-1B-$\epsilon0.1$-Le & 49.42\% & 67.14 & +2.34\% \\
  &
 4S-2B-$\epsilon0.1$-Le & 73.62\% & 67.08 & +5.86\% \\
  &
 8S-3B-$\epsilon1$-Le & 85.86\% & 67.42 & +9.90\% \\
  &
 16S-4B-$\epsilon5$-GaLe & 93.75\% & 64.88 & +10.94\% \\
\midrule
\multirow{6}{*}{MobilenetV2 a120}&
 - & - & 78.38 & - \\
  &
 2S-1B-$\epsilon0.1$-Le & 44.11\% & 78.38 & +2.34\% \\
  &
 4S-2B-$\epsilon0.1$-Le & 72.16\% & 78.38 & +7.42\% \\
  &
 8S-3B-$\epsilon1$-GaLe & 85.16\% & 78.38 & +19.53\% \\
  &
 8S-3B-$\epsilon1$-GaLe & 85.07\% & 78.38 & +19.53\% \\
  &
 16S-4B-$\epsilon5$-GaLe & 92.33\% & 78.24 & +34.38\% \\
\midrule
\multirow{6}{*}{MobilenetV2 a140}&
 - & - & 77.30 & - \\
  &
 2S-1B-$\epsilon0.1$-Le & 48.12\% & 77.28 & +2.34\% \\
  &
 4S-2B-$\epsilon0.1$-Le & 73.15\% & 77.28 & +5.86\% \\
  &
 8S-3B-$\epsilon1$-GaLe & 85.26\% & 77.32 & +14.32\% \\
  &
 16S-4B-$\epsilon5$-GaLe & 92.16\% & 77.28 & +27.34\% \\
\midrule
\multirow{4}{*}{MobilenetV3 Large a100}&
 - & - & 76.94 & - \\
  &
 2S-1B-$\epsilon0.1$-Le & 48.44\% & 76.94 & +2.34\% \\
  &
 4S-2B-$\epsilon0.1$-Le & 72.98\% & 76.94 & +5.86\% \\
  &
 16S-4B-$\epsilon5$-GaLe & 92.13\% & 76.56 & +16.30\% \\
\midrule
\multirow{6}{*}{MobilenetV3 L minimal a100}&
 - & - & 74.18 & - \\
  &
 2S-1B-$\epsilon0.1$-Le & 47.61\% & 74.18 & +2.34\% \\
  &
 4S-2B-$\epsilon0.1$-Le & 74.18\% & 74.18 & +5.86\% \\
  &
 8S-3B-$\epsilon1$-GaLe & 85.05\% & 74.16 & +15.36\% \\
  &
 16S-4B-$\epsilon5$-GaLe & 91.36\% & 74.18 & +32.23\% \\
  &
 32S-5B-$\epsilon10$-GaLe & 95.31\% & 72.56 & +39.06\% \\
 \midrule
 \multirow{5}{*}{MobilenetV4 Hybrid M}&
 - & - & 81.74 & - \\
  &
 2S-1B-$\epsilon0.1$-Le & 48.42\% & 81.74 & +1.56\% \\
  &
 4S-2B-$\epsilon0.1$-Le & 73.50\% & 81.74 & +11.72\% \\
  &
 8S-3B-$\epsilon50$-GaLe & 85.99\% & 80.66 & +14.6\% \\
  &
 16S-4B-$\epsilon20$-GaLe & 93.87\% & 75.66 & +37.50\% \\
\midrule
\multirow{3}{*}{MobilenetV4 Hybrid L}&
 - & - & 80.86 & - \\
  &
 2S-1B-$\epsilon0.1$-Le & 48.48\% & 80.86 & +1.56\% \\
  &
 4S-2B-$\epsilon0.1$-GaLe & 73.48\% & 78.61 & +18.72\% \\
\bottomrule
\end{tabular}
\caption{Additional results on timm models.}
\label{tab:additional_timm_1}
\end{table*}

\begin{table*}[tbp]
\centering
\begin{tabular}{llccc}
\toprule
Network & Configuration & RAM Reduction \% & Top-1 \% & Overhead \% \\
\midrule

\multirow{4}{*}{ResNet18}&
 - & - & 72.56 & - \\
  &
 2S-1B-$\epsilon0.1$-Le & 42.25\% & 72.54 & +14.06\% \\
  &
 4S-2B-$\epsilon0.1$-GaLe & 60.80\% & 72.50 & +32.03\% \\
  &
 16S-4B-$\epsilon5$-GaLe & 90.72\% & 64.12 & +13.28\% \\
\midrule
\multirow{4}{*}{ResNet50}&
 - & - & 80.06 & - \\
  &
 2S-1B-$\epsilon0.1$-Le & 45.17\% & 80.10 & +9.38\% \\
  &
 4S-2B-$\epsilon0.1$-Le & 65.61\% & 80.08 & +13.44\% \\
  &
 16S-4B-$\epsilon5$-GaLe & 92.05\% & 79.52 & +40.9\% \\
\midrule
\multirow{4}{*}{ResNet200}&
 - & - & 83.60 & - \\
  &
 2S-1B-$\epsilon0.1$-Le & 43.82\% & 83.60 & +12.50\% \\
  &
 4S-2B-$\epsilon0.1$-GaLe & 62.39\% & 83.60 & +40.62\% \\
  &
 16S-4B-$\epsilon20$-GaLe & 90.66\% & 81.94 & +45.12\% \\
\midrule
\multirow{5}{*}{EfficientNetB0}&
 - & - & 78.50 & - \\
  &
 2S-1B-$\epsilon0.1$-Le & 48.54\% & 78.54 & +2.34\% \\
  &
 4S-2B-$\epsilon0.1$-GaLe & 70.05\% & 78.54 & +2.81\% \\
  &
 8S-3B-$\epsilon1$-GaLe & 85.98\% & 78.54 & +14.58\% \\
  &
 16S-4B-$\epsilon5$-GaLe & 92.34\% & 78.30 & +21.88\% \\
\midrule
\multirow{5}{*}{EfficientNetB2}&
 - & - & 80.18 & - \\
  &
 2S-1B-$\epsilon0.1$-Le & 47.59\% & 80.18 & +3.91\% \\
  &
 4S-2B-$\epsilon0.1$-Le & 62.50\% & 80.14 & +18.75\% \\
  &
 8S-3B-$\epsilon1$-GaLe & 84.35\% & 80.12 & +19.53\% \\
  &
 16S-4B-$\epsilon5$-GaLe & 92.07\% & 79.46 & +23.83\% \\
\bottomrule
\multirow{5}{*}{EfficientViT B1}&
 - & - & 80.06 & - \\
  &
 2S-1B-$\epsilon0.1$-Le & 48.46\% & 80.10 & +2.34\% \\
  &
 4S-2B-$\epsilon0.1$-Le & 72.76\% & 80.08 & +5.86\% \\
  &
 8S-3B-$\epsilon1$-Le & 85.85\% & 80.00 & +12.50\% \\
  &
 16S-4B-$\epsilon5$-GaLe & 92.86\% & 78.56 & +12.70\% \\
\midrule
\multirow{5}{*}{EfficientViT B2}&
 - & - & 82.90 & - \\
  &
 2S-1B-$\epsilon0.1$-Le & 48.47\% & 82.88 & +2.34\% \\
  &
 4S-2B-$\epsilon0.1$-Le & 72.62\% & 82.92 & +7.42\% \\
  &
 8S-3B-$\epsilon1$-GaLe & 85.86\% & 82.90 & +15.10\% \\
  &
 16S-4B-$\epsilon5$-GaLe & 93.07\% & 82.72 & +20.12\% \\
\midrule
\multirow{4}{*}{ResNext101 32x8}&
 - & - & 82.88 & - \\
  &
 2S-1B-$\epsilon0.1$-Le & 43.71\% & 82.88 & +12.50\% \\
  &
 4S-2B-$\epsilon0.1$-GaLe & 62.44\% & 82.88 & +29.69\% \\
  &
 16S-4B-$\epsilon80$-GaLe & 93.69\% & 81.06 & +18.75\% \\
\midrule
\end{tabular}
\caption{Additional results on timm models.}
\label{tab:additional_timm_2}
\end{table*}

\end{document}